\documentclass[10pt,twocolumn,letterpaper]{article}

\usepackage{cvpr}
\usepackage{CJKutf8}

\definecolor{cvprblue}{rgb}{0.21,0.49,0.74}
\usepackage[pagebackref,breaklinks,colorlinks,allcolors=cvprblue]{hyperref}

\def\paperID{*****}
\def\confName{CVPR}
\def\confYear{2026}

\title{Local Brushstroke Quality Assessment via Vision-Language Feedback}

\author{Mio Mitamura\textsuperscript{1}\\
$^1$Tokyo Institute of Science High School
\and
Hirokatsu Kataoka\textsuperscript{2,3}\\
$^2$National Institute of Advanced Industrial Science and Technology (AIST)\\
$^3$Visual Geometry Group, University of Oxford
}

\begin{document}
\maketitle

\begin{abstract}
This paper investigates whether multimodal LLMs can evaluate local brushstroke quality in calligraphy and generate educationally useful natural language feedback. We construct an evaluation framework in which three multimodal LLMs (GPT-4o~\cite{openai2023gpt4}, Claude Sonnet 4~\cite{claude2024}, and Gemini 2.5 Flash~\cite{gemini2023}) assess before-after image pairs of calligraphic works using a five-point ordinal scale, and compare their outputs against scores assigned by three expert calligraphers. We additionally examine a Retrieval-Augmented Generation (RAG)~\cite{lewis2020rag} variant of Claude as a preliminary condition. Results show that all models achieve useful levels of absolute score accuracy (MAE), with GPT-4o performing best (MAE = 0.885). However, none of the models produce statistically significant overall rank correlations with human experts (Kendall's $\tau$). Vocabulary analysis of generated rationales reveals characteristic evaluative biases in each model, and RAG is shown to improve rank correlation while worsening absolute accuracy, constituting an important negative result for text-based rule injection.
\end{abstract}


\section{Introduction}

Calligraphy is a traditional art form deeply rooted in Chinese and Japanese culture. Unlike ordinary writing, calligraphy demands not only accurate character shape but also mastery of brushstroke techniques such as \textit{tome} (stop), \textit{hane} (hook), and \textit{harai} (sweep). These techniques are fundamental to calligraphic quality, yet their assessment has traditionally relied on skilled instructors, making objective and automated evaluation extremely difficult.

\begin{figure}[t]
  \centering
  \includegraphics[width=\linewidth]{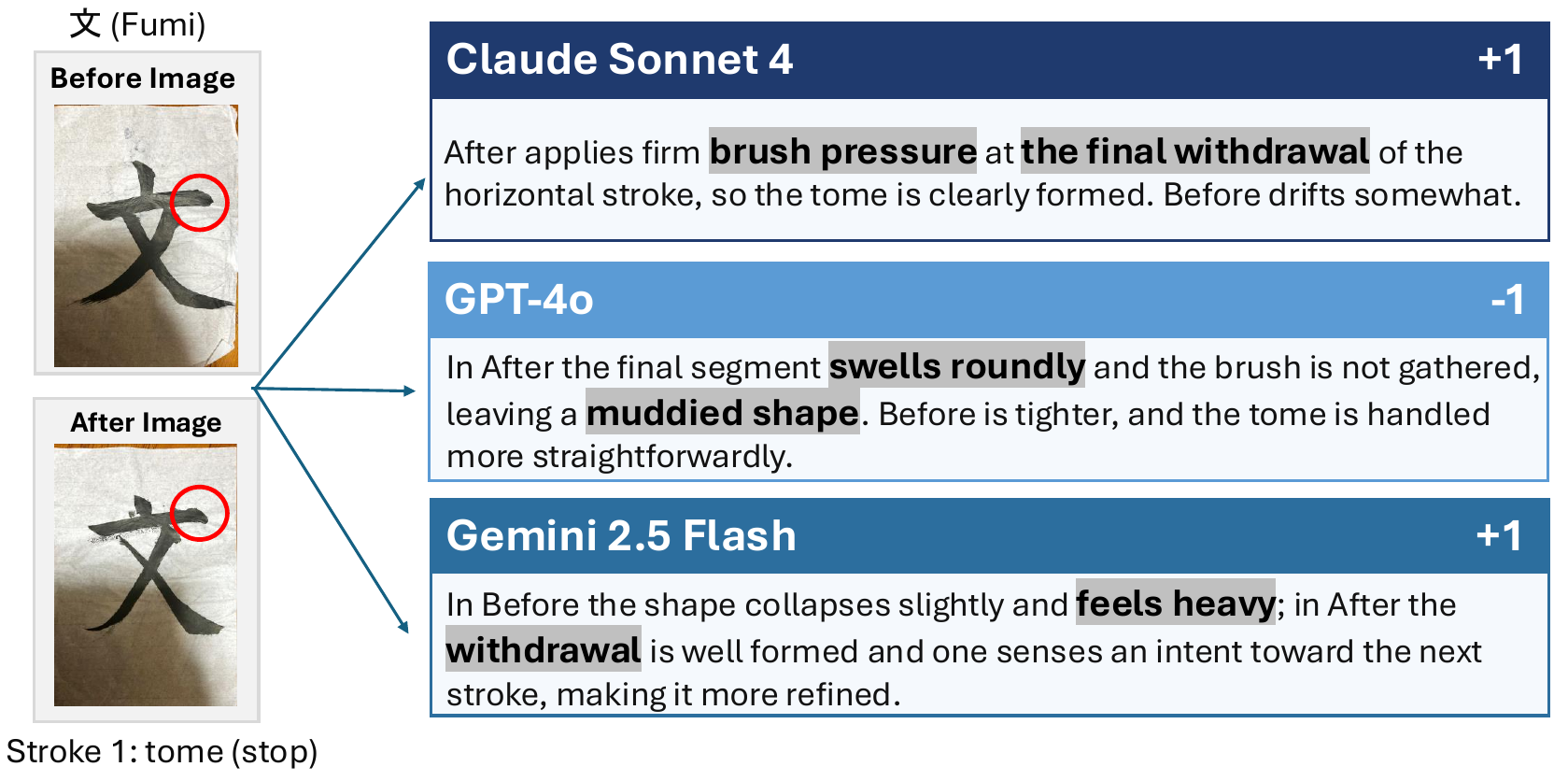}
  \caption{Three multimodal LLMs evaluate the same local stroke (\begin{CJK}{UTF8}{min}文\end{CJK}, Stroke 1: \textit{tome}). Scores range from $-2$ (before clearly better) to $+2$ (after clearly better). Claude and Gemini both score $+1$, while GPT-4o scores $-1$ on the identical stroke. Highlighted spans mark the terms in which each model grounds its judgment: Claude cites brush-execution terms (\textit{brush pressure}, \textit{final withdrawal}), whereas GPT-4o cites impressionistic ones (\textit{swells roundly}, \textit{muddied shape}). The models disagree not only in score but in the kind of language they use.}
  \label{fig:fig1}
\end{figure}

Recent advances in AI-driven automated assessment have transformed several educational domains, including foreign language learning and mathematics~\cite{ramesh2022aes}. However, calligraphy presents unique challenges: in addition to static character form, evaluation must capture dynamic qualities such as brush pressure and the rhythm of stroke execution---aspects that existing image-based methods are ill-equipped to address.

Prior work on automated calligraphy assessment has primarily focused on static features such as character shape and structural regularity~\cite{yoshida2023,li2023}. The importance of dynamic brushstroke characteristics has also been noted~\cite{xu2018,yuan2025}. However, existing approaches generally output only a numerical score, without providing the stroke-specific instructional feedback that learners and educators require. Furthermore, inferring the dynamic qualities of a brushstroke from a static two-dimensional image remains a fundamental open challenge, paralleling challenges in action quality assessment~\cite{aqa_survey2025}.

Multimodal LLMs are a class of models that encode visual inputs through an image encoder and inject the resulting visual features into a Large Language Model (LLM), enabling the generation of natural language descriptions and reasoning grounded in visual content~\cite{openai2023gpt4,gemini2023,claude2024}. This architecture offers two properties that are potentially relevant to calligraphy assessment. First, the visual encoder may capture fine-grained spatial features of brushstrokes, such as the tapering of a \textit{harai} sweep or the curvature at the tip of a \textit{hane} hook, that are difficult to quantify with hand-crafted metrics, though the extent to which current models successfully do so remains an open question. Second, the language generation capability of the LLM component enables the model to produce stroke-specific natural language feedback, going beyond a scalar score to articulate the qualitative characteristics of brushstroke technique. These two properties motivate the investigation of multimodal LLMs as candidates for automated calligraphy assessment and instructional feedback generation.

Recent work has applied multimodal LLMs to calligraphy-related tasks. Luo et al.~\cite{calliReader2025} proposed CalliReader, a multimodal LLM-based framework for Chinese Calligraphy Contextualization (CC$^2$), demonstrating that multimodal LLMs can recognize and interpret calligraphic content at a level surpassing human experts in character recognition tasks. However, their focus is on text content recognition rather than brushstroke technique evaluation. The capability of multimodal LLMs to assess fine-grained local stroke quality and generate stroke-specific instructional feedback remains unexplored.

The broader landscape of multimodal LLMs has advanced rapidly. Zhang et al.~\cite{zhang2023gpt4v_eval} showed that GPT-4V can serve as a generalist evaluator for vision-language tasks, showing promising agreement with human judgment across diverse settings. However, recent work has also revealed that LLMs used as evaluators exhibit systematic biases~\cite{ye2024biases,zheng2023judging}, tending to collapse multi-dimensional judgment into dominant single perspectives. These findings motivate our investigation into whether such biases also manifest in the specialized domain of calligraphy assessment.

Compounding these technical difficulties is a broader societal trend: the number of calligraphy instructors and learners is declining. Between 2012 and 2021, the number of employees in calligraphy instruction businesses in Japan decreased by 33.1\%~\cite{census2012,census2016,census2021}. AI-assisted evaluation therefore holds promise as a means of expanding learning opportunities and maintaining educational quality.

In this work, we investigate whether multimodal LLMs can evaluate local brushstroke quality and generate educationally useful natural language feedback. Beyond producing numerical scores, we explore the extent to which multimodal LLMs can articulate, in natural language, the strengths and weaknesses of specific brushstroke techniques. We additionally conduct a preliminary investigation into Retrieval-Augmented Generation (RAG)~\cite{lewis2020rag}, examining its effectiveness and limitations when expert evaluation criteria are provided as external knowledge. As illustrated in \cref{fig:fig1}, multimodal LLMs evaluating the same local stroke can disagree not only in their scores but in the kind of language they use to justify them.

The contributions of this paper are threefold. First, we provide a quantitative evaluation of multimodal LLM-based local brushstroke assessment in calligraphy. Second, we analyze the natural language feedback generated by each model to clarify the differences in evaluation perspective between AI systems and human experts. Third, as a preliminary analysis, we examine the effects and limitations of RAG and demonstrate how providing external evaluation criteria alters model behavior.

\section{Method}

\begin{figure*}[t]
  \centering
  \includegraphics[width=0.90\linewidth]{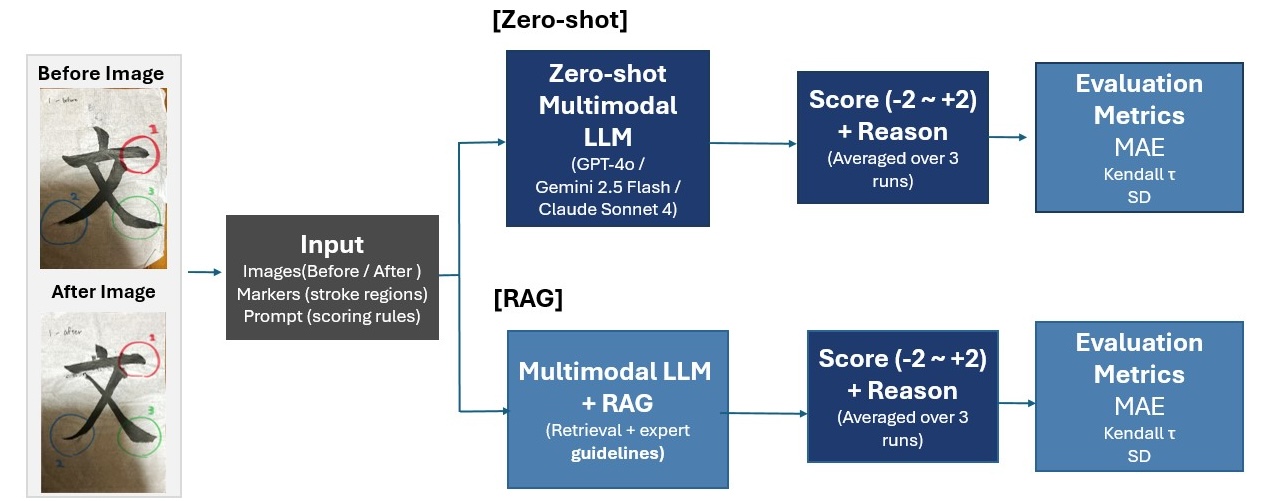}
  \vspace{-0pt}
  \caption{Overview of the proposed evaluation pipeline. Before and After calligraphy images with stroke markers are input to zero-shot multimodal LLMs and multimodal LLM+RAG, each outputting a score ($-2$ to $+2$) and natural language rationale.}
  \label{fig:pipeline}
\end{figure*}

We constructed an evaluation framework in which multimodal LLMs assess brushstroke quality in calligraphy works. The framework takes as input a before image and an after image of the same calligraphic work, along with visual markers indicating the local stroke region of interest, and produces a comparative assessment of brushstroke technique quality (see \cref{fig:pipeline}).

For each work, three stroke regions exhibiting salient brushstroke characteristics were selected and annotated with markers. Each region was associated with a specific stroke type: \textit{tome} (stop), \textit{hane} (hook), or \textit{harai} (sweep, either left or right). The model was asked to judge which of the two images (before or after images) demonstrated superior technique, using a five-point ordinal scale: $-2$ (before clearly better), $-1$ (before slightly better), $0$ (equivalent), $+1$ (after slightly better), $+2$ (after clearly better). For each annotated region, the model output both a numerical score and a natural language rationale, enabling both quantitative measurement and qualitative analysis.

The models evaluated were Claude Sonnet 4~\cite{claude2024}, Gemini 2.5 Flash~\cite{gemini2023}, and GPT-4o~\cite{openai2023gpt4}, all assessed in a zero-shot setting without an additional training. In addition, a RAG-augmented~\cite{lewis2020rag} variant of Claude was examined as a preliminary condition. In the RAG condition, a knowledge base was constructed from expert evaluation examples (comprising calligraphy images, assigned scores, and written rationales) together with stroke-specific evaluation guidelines.

All models received identical prompt instructions specifying the target character, the stroke region under evaluation, the five-point scoring rubric, and the required output format. To avoid anchoring bias, models were explicitly instructed not to assume that either the before or after image was superior and to evaluate solely on technical grounds. To mitigate score distribution skew, conditions for assigning extreme scores ($\pm2$) and the neutral score ($0$) were explicitly stated.

To reduce output variability, each model was run three independent times per stroke region, and the mean score across runs was taken as the final evaluation. Performance was assessed against the mean scores of three human experts using Mean Absolute Error (MAE), Kendall's $\tau$ rank correlation, and standard deviation (SD). Generated natural language rationales were further subjected to vocabulary analysis to characterize the evaluative perspectives of each model.

\section{Experiment}

\subsection{Experimental Setup}

\noindent\textbf{Dataset.}
The dataset consisted of calligraphy images collected during a user study conducted at a school cultural festival. Three Japanese kanji characters were included: \begin{CJK}{UTF8}{min}文\end{CJK} (Fumi), \begin{CJK}{UTF8}{min}化\end{CJK} (Ka), and \begin{CJK}{UTF8}{min}祭\end{CJK} (Matsuri). Each sample comprised a before-after image pair, yielding 45 works in total. For each work, three stroke regions showing pronounced brushstroke characteristics were selected for evaluation.

The stroke regions evaluated per character were as follows. For \begin{CJK}{UTF8}{min}文\end{CJK}: Stroke 1 (\textit{tome}/stop), Stroke 2 (\textit{hidari-harai}/left sweep), Stroke 3 (\textit{migi-harai}/right sweep). For \begin{CJK}{UTF8}{min}化\end{CJK}: Stroke 1 (\textit{hidari-harai}/left sweep), Stroke 2 (\textit{tome}/stop), Stroke 3 (\textit{hane}/hook). For \begin{CJK}{UTF8}{min}祭\end{CJK}: Stroke 1 (\textit{hidari-harai}/left sweep), Stroke 2 (\textit{migi-harai}/right sweep), Stroke 3 (\textit{hane}/hook).

For the RAG condition, 35 of the 45 works were used to construct the knowledge base, and the remaining 10 works were reserved as a held-out test set.

\noindent\textbf{Ground Truth Labels.}
Ground truth labels were defined as the mean score assigned by three expert calligraphers, each with at least ten years of practice. Rather than evaluating overall character shape, experts focused exclusively on local brushstroke technique in each annotated region. The evaluation criteria were as follows:
\begin{itemize}
    \item \textit{hane}: whether the brush is firmly stopped before being lifted with a controlled release of pressure.
    \item \textit{migi-harai}: whether the brush flows smoothly downward and to the right after stopping, tapering gradually.
    \item \textit{hidari-harai}: whether the brush flows smoothly downward and to the left after stopping, tapering gradually.
    \item \textit{tome}: whether the brush stops firmly to form a stable, well-defined endpoint.
\end{itemize}

Each evaluation compared the before and after images using the five-point scale ($-2$ to $+2$). The mean of the three experts' scores served as the ground truth (\textit{human\_avg}). In addition to numerical scores, written rationales were collected for subsequent qualitative analysis.

\noindent\textbf{Models Evaluated.}
Three multimodal LLMs were evaluated: GPT-4o~\cite{openai2023gpt4}, Gemini 2.5 Flash~\cite{gemini2023}, and Claude Sonnet 4~\cite{claude2024} (zero-shot). A RAG-augmented~\cite{lewis2020rag} Claude condition was additionally examined as a preliminary study. The RAG model was evaluated only on the held-out test set (10 works, 30 strokes). All models received identical input conditions, and each stroke region was evaluated in three independent runs.

\noindent\textbf{Evaluation Metrics.}
Quantitative metrics comprised: MAE (Mean Absolute Error); Kendall's $\tau$ rank correlation; and SD (Standard Deviation). For qualitative analysis, vocabulary was coded into four categories: \textit{Static}, \textit{Dynamic-Action}, \textit{Dynamic-Impression}, and \textit{Quality}.

\subsection{Main Results on 45 Works}

\noindent\textbf{Quantitative Evaluation.}
\cref{tab:results} presents the quantitative results. GPT-4o achieved the best MAE (0.885), followed by Claude (0.902). Claude achieved the lowest SD (0.481). For Kendall's $\tau$, GPT-4o achieved $\tau = 0.093$, Claude $\tau = 0.053$, and Gemini $\tau = -0.025$. None reached statistical significance (all $p > 0.05$), suggesting that reliable rank ordering remains a substantially harder task than absolute scoring.

\begin{table}[htbp]
  \caption{Quantitative evaluation results of multimodal LLMs vs.\ human expert scores. GPT-4o achieves the best MAE (0.885), Claude shows the most stable output (SD: 0.481). None achieve statistically significant overall rank correlation with human experts (all $p > 0.05$).}
  \label{tab:results}
  \centering
  \includegraphics[width=\linewidth]{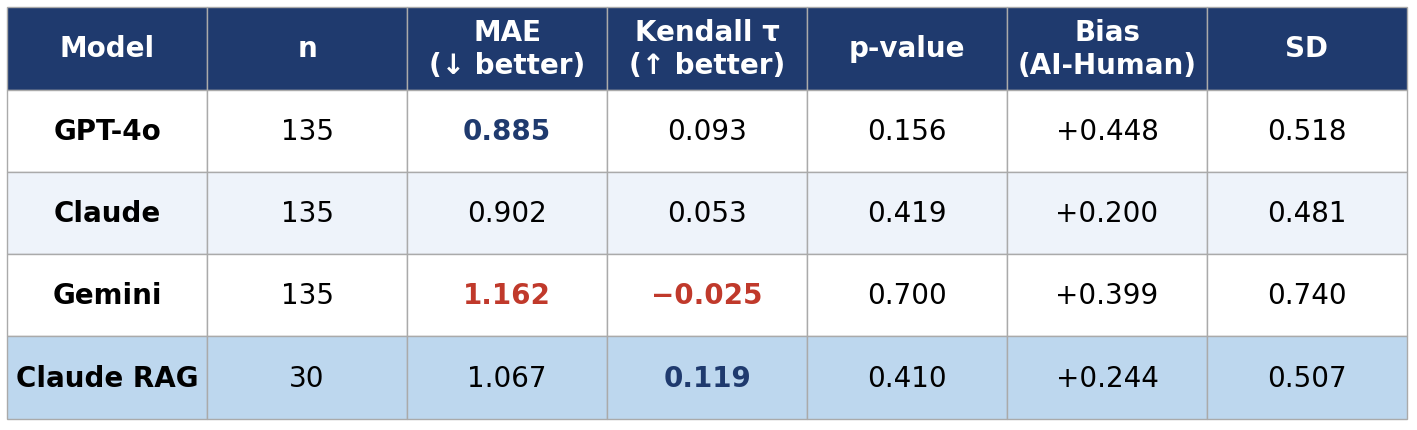}
\end{table}

\begin{figure}[htbp]
  \centering
  \includegraphics[width=\linewidth]{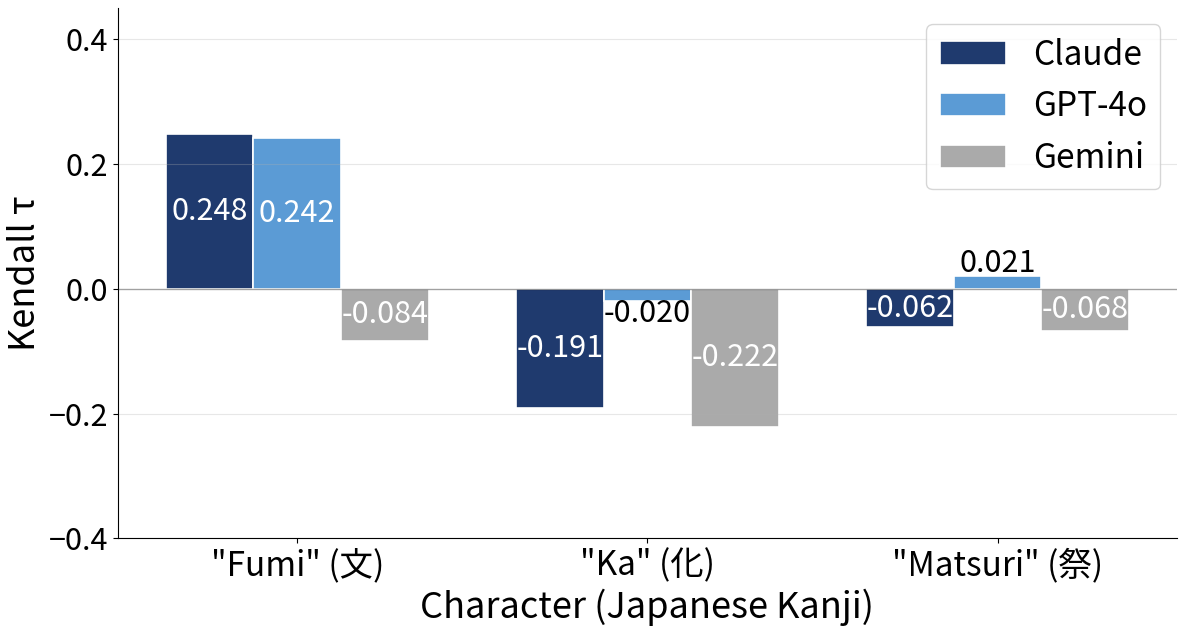}
  \caption{Kendall $\tau$ by character type. Significant correlations ($p < 0.05$) appear only for ``Fumi'' (\begin{CJK}{UTF8}{min}文\end{CJK}), the simplest character. Complex characters show no significant correlations, suggesting multimodal LLMs cannot ignore overall visual complexity.}
  \label{fig:kendall_char}
\end{figure}

\noindent\textbf{Analysis by Character and Stroke Type.}
\cref{fig:kendall_char} shows Kendall's $\tau$ by character. Claude ($\tau = 0.248$, $p = 0.025$) and GPT-4o ($\tau = 0.242$, $p = 0.027$) achieved significant correlations only for \begin{CJK}{UTF8}{min}文\end{CJK} (Fumi, simple 4-stroke). Both Claude and Gemini showed negative correlations for \begin{CJK}{UTF8}{min}化\end{CJK} ($\tau = -0.191$ and $-0.222$), suggesting multimodal LLMs are implicitly influenced by overall structural complexity.

\begin{figure}[htbp]
  \centering
  \includegraphics[width=\linewidth]{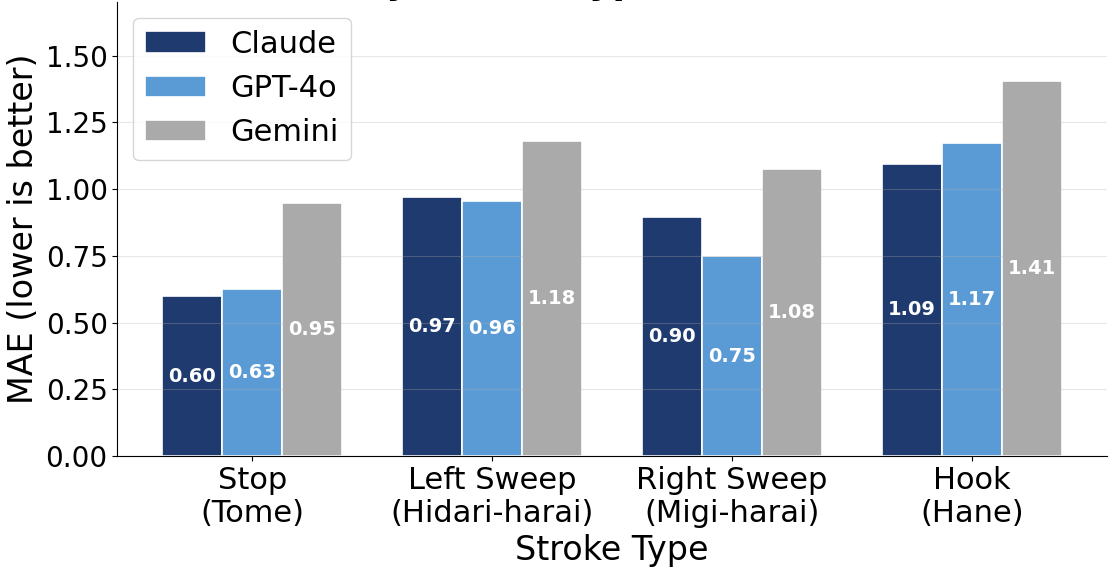}
  \caption{MAE by stroke type. All models show lowest error for Tome (Stop) and highest error for Hane (Hook), which requires inferring a multi-phase 3D brushstroke process from a static 2D image.}
  \label{fig:mae_stroke}
\end{figure}

\cref{fig:mae_stroke} shows MAE by stroke type. \textit{Tome} (stop) yielded the lowest MAE (Claude: 0.600, GPT-4o: 0.627), while \textit{hane} (hook) produced the largest errors (Claude: 1.095, GPT-4o: 1.173, Gemini: 1.405). The \textit{hane} stroke demands inferring a multi-phase 3D process from a static 2D image, reflecting inherent limits consistent with challenges in action quality assessment~\cite{aqa_survey2025}.

\begin{figure}[htbp]
  \centering
  \includegraphics[width=\linewidth]{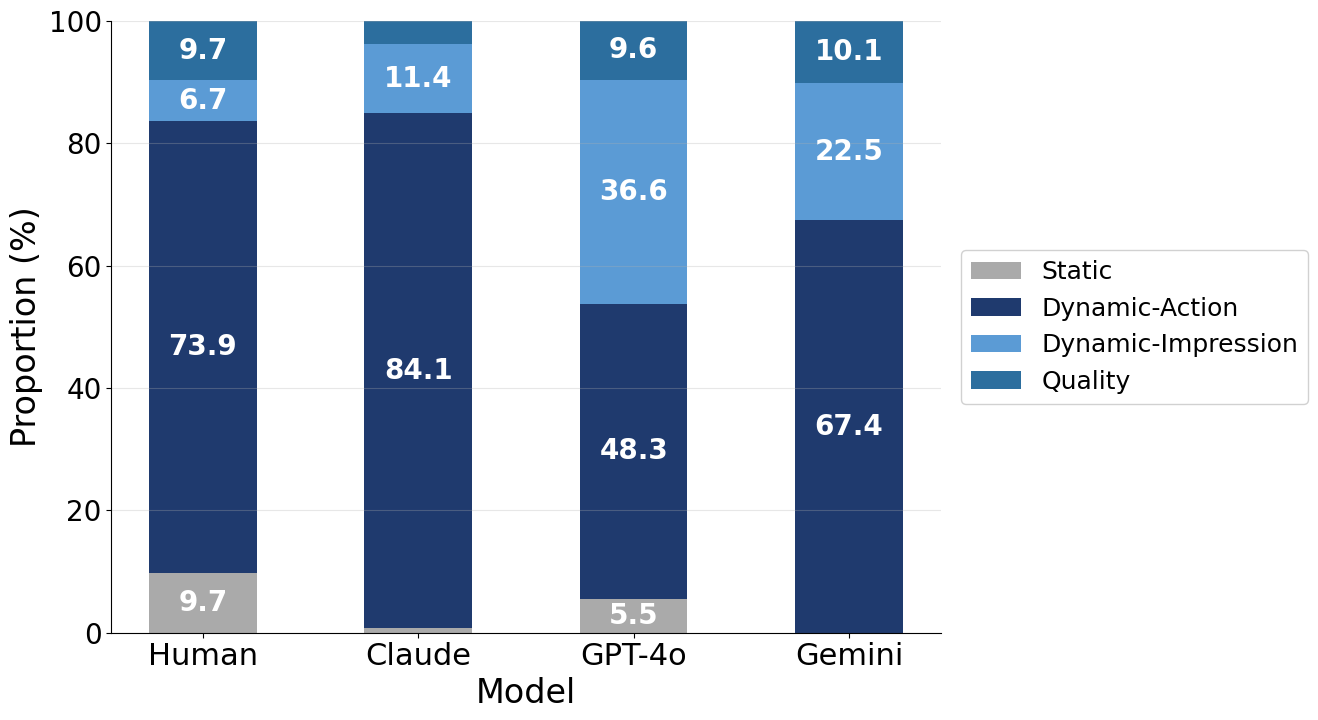}
  \caption{Vocabulary distribution by model. Human evaluators use all four categories, while each AI model exhibits a characteristic bias: Claude collapses into dynamic-action, and both GPT-4o and Gemini shift toward dynamic-impression. Among the models, Gemini shows the highest Quality proportion. \textit{Static} vocabulary is virtually absent from all AI outputs.}
  \label{fig:vocab}
\end{figure}

\noindent\textbf{Analysis of Natural Language Feedback.}
\cref{fig:vocab} shows vocabulary distributions. Human evaluators used all four categories (Dynamic-Action: 73.9\%, Static: 9.7\%). By contrast, AI models exhibited characteristic biases: Claude collapsed into Dynamic-Action (84.1\%), while GPT-4o and Gemini both shifted toward Dynamic-Impression (36.6\% and 22.5\%, vs.\ 6.7\% for humans). Among the models, Gemini showed the highest proportion of Quality vocabulary (10.1\%). \textit{Static} vocabulary was virtually absent in all AI outputs (at most 5.5\% for GPT-4o). These characteristic biases are consistent with known tendencies of LLMs as evaluators~\cite{ye2024biases,zheng2023judging}, and with prior observations that automated scoring systems tend to collapse multi-dimensional human judgment~\cite{ramesh2022aes,misgna2024survey}.

\subsection{Preliminary Analysis of RAG-Augmented Evaluation}

\noindent\textbf{Experimental Setup.}
We compared Claude zero-shot with Claude+RAG~\cite{lewis2020rag} on 30 held-out strokes, providing stroke-specific checklist items as external knowledge.

\begin{figure}[htbp]
  \centering
  \includegraphics[width=\linewidth]{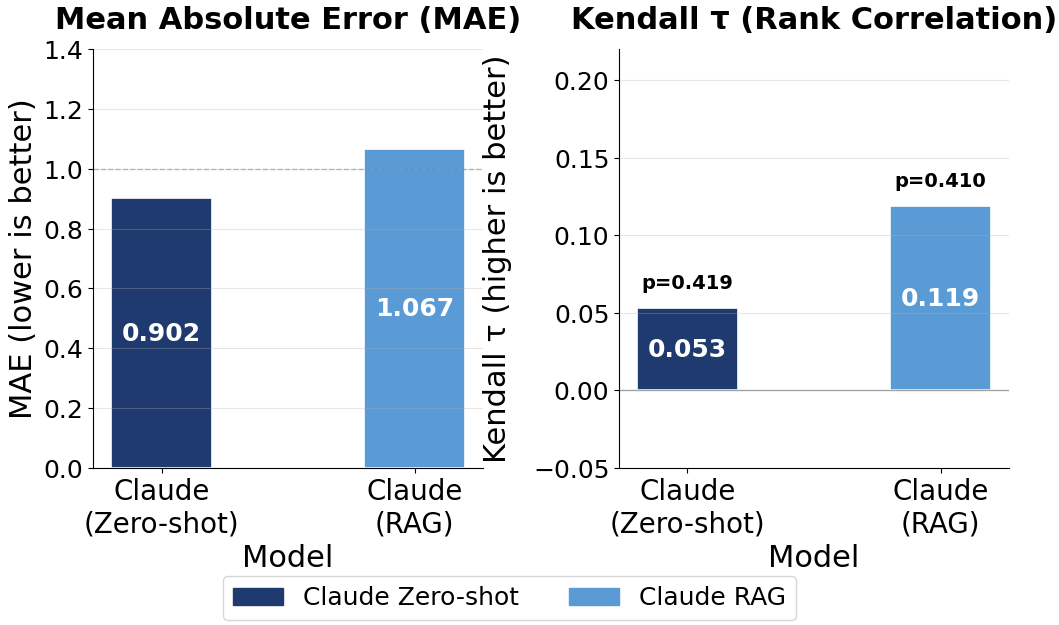}
  \caption{Claude Zero-shot vs.\ Claude RAG ($n=30$). RAG improves rank correlation ($\tau$: $0.053 \to 0.119$) while increasing absolute error (MAE: $0.902 \to 1.067$), revealing a fundamental trade-off.}
  \label{fig:rag_compare}
\end{figure}

\noindent\textbf{Quantitative Comparison.}
The results (\cref{fig:rag_compare}) revealed a trade-off: MAE increased from 0.902 to 1.067, while Kendall's $\tau$ improved from 0.053 to 0.119. Explicit criteria stabilize relative comparisons but introduce divergence from holistic human scoring.

\begin{figure}[htbp]
  \centering
  \includegraphics[width=\linewidth]{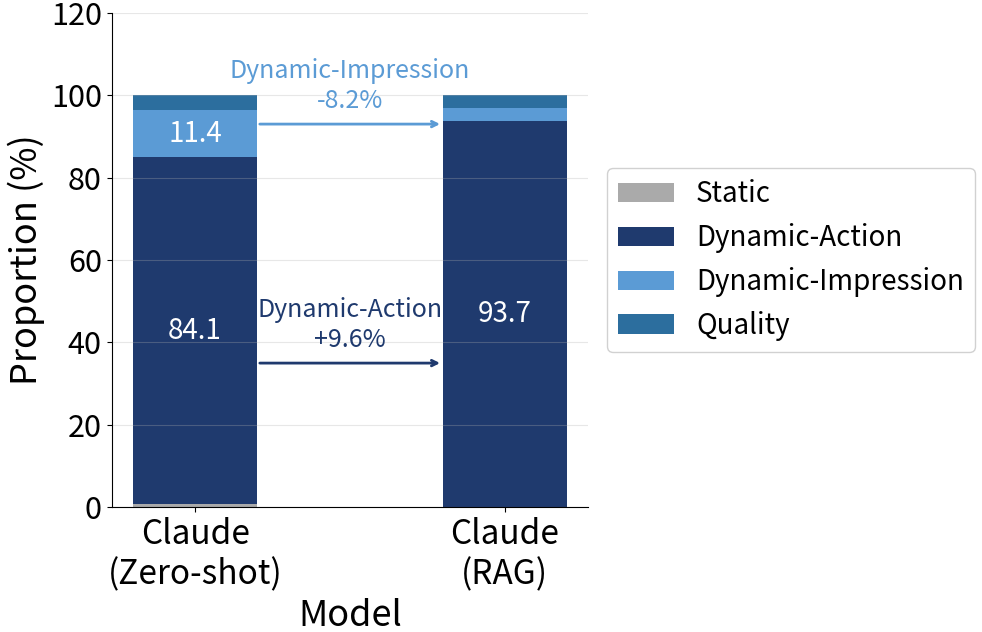}
  \caption{Vocabulary: Claude Zero-shot vs.\ Claude RAG. RAG concentrates language toward dynamic-action ($84.1\% \to 93.7\%$), causing near-complete disappearance of Static, Quality, and dynamic-impression terms.}
  \label{fig:vocab_rag}
\end{figure}

\begin{figure}[htbp]
  \centering
  \includegraphics[width=\linewidth]{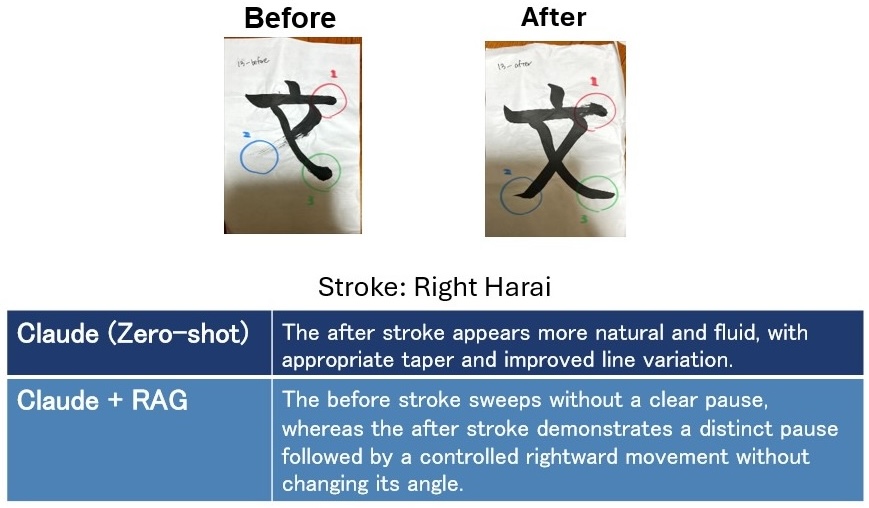}
  \caption{Feedback differences (\begin{CJK}{UTF8}{min}文\end{CJK}, right \textit{harai}). Zero-shot generates holistic aesthetic feedback, while RAG over-concentrates on dynamic-action vocabulary, losing the multi-dimensional balance that human experts naturally integrate.}
  \label{fig:rag_example}
\end{figure}

\noindent\textbf{Changes in Feedback Content.}
In the RAG condition, 93.7\% of vocabulary was dynamic-action (up from 84.1\%), and \textit{Static} vocabulary disappeared entirely (\cref{fig:vocab_rag}). \cref{fig:rag_example} illustrates the qualitative shift: zero-shot output integrates aesthetic language, while RAG output reduces to mechanical checklist-checking. These findings indicate that text-based rule injection does not induce human-like holistic evaluation, and more sophisticated approaches are needed.

\section{Conclusion}

\subsection{Summary}
Experiments on 45 works showed that GPT-4o~\cite{openai2023gpt4}, Claude~\cite{claude2024}, and Gemini~\cite{gemini2023} achieve useful absolute accuracy (best MAE = 0.885), but none produce statistically significant overall rank correlations. Vocabulary analysis revealed characteristic evaluative biases in each model, consistent with known LLM evaluation biases~\cite{ye2024biases,zheng2023judging}. RAG~\cite{lewis2020rag} constitutes an important negative result: rule injection does not induce human-like holistic evaluation. This contrasts with calligraphy content recognition~\cite{calliReader2025}, where multimodal LLMs surpass humans, suggesting that brushstroke quality evaluation poses fundamentally different challenges.

\subsection{Implications}
Two structural limitations emerge. First, multimodal LLMs cannot fully ignore overall visual complexity when instructed to evaluate local regions. Second, despite multimodal LLMs' ability to inject visual features into an LLM and generate natural language descriptions, the resulting evaluations collapse multi-dimensional human judgment into a single dominant perspective, consistent with challenges in action quality assessment~\cite{aqa_survey2025} and automated scoring~\cite{ramesh2022aes}.

\subsection{Future Work}
Future work includes fine-tuning on expert evaluation data, developing multimodal attention mechanisms capable of independently capturing local brushstroke process and static form, larger-scale RAG~\cite{lewis2020rag} validation, and practical educational validation with calligraphy learners.

{\small
\paragraph{Acknowledgment.}
The authors thank the student participants who contributed calligraphy works during the school cultural festival user study, and the expert calligraphers who provided ground truth evaluations. This work was conducted as part of the Experts in Information Science Program of the National Institute of Informatics (NII), which also provided financial support for this research.
}

{\small
\bibliographystyle{ieeenat_fullname}

}

\end{document}